%% file: main.tex
\documentclass{article} 
\usepackage{iclr2025_conference,times}

\input{math_commands.tex}

\usepackage{hyperref}
\usepackage{url}
\usepackage{subcaption}
\usepackage{booktabs}
\usepackage{multirow}
\usepackage{makecell}
\usepackage{amssymb}
\usepackage{graphicx}
\usepackage{newfloat}
\usepackage{listings}
\usepackage{caption}
\usepackage{adjustbox}
\usepackage{enumitem}
\usepackage{cleveref}
\usepackage{fontawesome5}

\title{LipShiFT: A Certifiably Robust Shift-based \\ Vision Transformer}

\author{Rohan Menon$^1$, Nicola Franco$^{2}$, Stephan Günnemann$^1$ \\
$^1$ Technical Univ. of Munich, School of Computation and Information Technology\\
$^2$ Fraunhofer Institute for Cognitive Systems IKS, Munich, Germany
}

\iclrfinalcopy 
\begin{document}

\maketitle

\begin{abstract}
Deriving tight Lipschitz bounds for transformer-based architectures presents a significant challenge. The large input sizes and high-dimensional attention modules typically prove to be crucial bottlenecks during the training process and leads to sub-optimal results. Our research highlights practical constraints of these methods in vision tasks. We find that Lipschitz-based margin training acts as a strong regularizer while restricting weights in successive layers of the model. Focusing on a Lipschitz continuous variant of the ShiftViT model, we address significant training challenges for transformer-based architectures under norm-constrained input setting.
We provide an upper bound estimate for the Lipschitz constants of this model using the $l_2$ norm on common image classification datasets. 
Ultimately, we demonstrate that our method scales to larger models and advances the state-of-the-art in certified robustness for transformer-based architectures\footnote{\faGithub \space\url{https://github.com/RohanMenon/LipShiFT}}.
\end{abstract}

\section{Introduction}\label{sec:introduction}

Vision transformers have been extremely versatile and considered as foundational breakthroughs in deep learning (DL)~\citep{dosovitskiy2021imageworth16x16words}. For applications in the field of computer vision they can easily expand to multiple domains and varied number of tasks such as classification~\citep{dosovitskiy2021imageworth16x16words, touvron2021trainingdataefficientimagetransformers}, segmentation~\citep{ye2019cross} and object detection~\citep{carion2020endtoendobjectdetectiontransformers, zhu2021deformabledetrdeformabletransformers}. Although compared to other popular vision architectures such as residual networks (ResNets)~\citep{he2016deep} and convolutional networks (ConvNets)~\citep{lecun1989backpropagation}, the effect of adversarial attacks on transformer-based models has been studied in limited capacity~\citep{shao2021adversarial}. 
An adversarial attack is defined as injecting noise to an input such that it disrupts the model's decision making process~\citep{akhtar2018threat}. 
This poses significant risks of severe consequences when models are deployed in real-world safety-critical contexts.

Formal verification of DL models is crucial for guaranteeing their reliability and compliance with standards in safety-critical applications, such as autonomous driving or healthcare~\citep{urban2021review}. 
However, the current approaches for formal verification have been struggling to efficiently scale to large architectures~\citep{konig2024critically}. 
Transformer-based architectures, being one of them, are not researched as frequently in the context of formal verification although their wide range of applicability in vision and natural language processing is very well known. 

In this paper, we focus on applying adversarial training to a transformer-based architecture to obtain provable Lipschitz-based robustness certificates.  
We summarize our contributions here:

\begin{enumerate}[label=\roman*.] 
    \item Introduce \textbf{LipShiFT}, a transformer-based Lipschitz continuous architecture with tight Lipschitz bounds.
    \item Provide robustness guarantees under $l_2$-norm for a certification radius $\epsilon = 36/255$.
    \item Evaluation of empirical robustness with AutoAttack~\citep{croce2020reliable}.
 \end{enumerate}

\section{Background}\label{sec:background}

The introduction of the \textit{transformer} architecture has paved the way to many notable developments in DL applications~\citep{vaswani2023attentionneed}. 
Notably in computer vision, it was adapted to give rise to Vision transformer (ViT)~\citep{dosovitskiy2021imageworth16x16words} which featured an N x N input image split into multiple non-overlapping patches and converted intro vector embeddings to conform to a sequence based input to the transformer. This input passes through many layers, one of which being the attention mechanism that was introduced to surpass the shortcomings of the Long-Short Term Memory (LSTM)~\citep{hochreiter1997long} and Recurrent Recurrent Neural Networks (RNN)~\citep{schuster1997bidirectional} models to capture long-term dependencies in the input. LSTM's traditionally suffered with the issue of vanishing gradients as the input sequence increases in length overtime. The original attention mechanism, introduced in 2017 paper by~\cite{vaswani2023attentionneed} was a Dot Product (DP) based mechanism that was effectively able to capture semantic meaning while being able to produce stable training.

\paragraph{Lipschitz continuity}
Given inputs $\mathbf{x}, \mathbf{y} \in \mathbb{R}^{N\times N}$ and function $\textbf{f} \colon \mathbb{R}^{N\times N} \to \mathbb{R}^M$. 
Formally, $\textbf{f}(\mathbf{x})$ is Lipschitz continuous, whose output bounded as follows:
\begin{equation} \label{eq: lipschitz-condition}
    || \textbf{f}(\mathbf{y}) - \textbf{f}(\mathbf{x}) || \leq L || \mathbf{y} - \mathbf{x} ||, L \geq 0,
\end{equation}
where $L$ is defined as the Lipschitz constant of the function $\textbf{f}$ and $|| \cdot ||$ is an $L_p$-norm.

The original transformer architecture, which relies on a dot-product mechanism, is not Lipschitz continuous. 
To address this, \citet{kim2021lipschitz} introduced an $L_2$-based self-attention transformer that satisfies the Lipschitz condition. 
However, this approach comes with its own challenges. The computation of the Lipschitz constant for $L_2$ attention is dependent on the input size, which makes it difficult to obtain a tight upper bound in practice for transformer-based architectures~\citep{hu2023scaling}.

\paragraph{ShiftViT}
Thus, we consider the ShiftViT~\citep{wang2022shiftoperationmeetsvision} which implements a partial channel shift operation which helps simulate the self-attention module and provide similar performance on ImageNet~\citep{imagenet_cvpr09}. 
This architectural change essentially provides us with a model that is lightweight and still can perform well compared to its attention-based alternatives. 
Providing robustness guarantees and certificates for transformer-based architectures has been a challenge in the past. To achieve a robust LipShiFT model, we utilize the Lipschitz margin loss function introduced by \citet{hu2023scaling} known as Efficient Margin Maximization (EMMA) Loss. It uses the model Lipschitz constant to maximize the class decision boundary for all classes with respect to the predicted class. This way, even if the top threatening class label changes frequently, it does not negatively affect the models epoch-wise robustness. Additionally, EMMA also uses a variable $\epsilon$ certification radius during training which helps it balance both the clean as well as verified accuracy while performing loss optimization.

\section{Related Work}\label{sec:related_work}

In this section, we briefly describe some of the most notable works which focus on Lipschitz continuous architectures and some that use Lipschitz margin training to guarantee model robustness. 
We focus mainly on applications in the vision domain~\citep{prach20231lipschitzlayerscomparedmemory}.

Local-Lipschitz (Local-Lip)~\citep{huang21local_lipschitz} introduces an efficient algorithm to compute the local Lipschitz constant of neural networks by computing the tight upper bound norm of the Clarke Jacobian via linear bound propagation. This method allows for the handling of larger and practical models that previous methods could not efficiently process. Although their results are applicable on a wide range of datasets like MNIST~\citep{deng2012mnist}, CIFAR10~\citep{cifar10} and Tiny ImageNet~\citep{Le2015TinyIV} for a certification radius of $\epsilon=36/255$, they report results only on fully convolutional networks and not on other scalable architectures such as transformers.
 
Globally Robust (GloRo) Networks~\citep{leino2021globallyrobustneuralnetworks} describes a new architecture such that it computes the global Lipschitz bound of the model which is tighter as compared to Local-Lip. GloRo primarily provides robustness guarantees by training the GloroNet using TRADES loss which is updated to utilize the model Lipschitz constant to maximize the decision boundary of the predicted class $i$ with respect to the top threatening class $j$. In principal, \citet{leino2021globallyrobustneuralnetworks} introduces a new $\perp \in \mathbb{Z}$ class label. This label is always $0$ in the one-hot encoded version of class labels, signifying a non-robust sample and assigned only when the model wrongly classifies the sample compared to the truth label. Based on their method, Gloro Networks achieve much better results compared to Local-Lip. Although they primarily also focus on fully convolutional network architectures.

\paragraph{Vision transformers}
A Certifiable Vision Transformer (CertViT)~\citep{gupta2023certvit} has been introduced with a two-step certification algorithm to provide robust certification for pre-trained vision transformers. In their paper, \citet{gupta2023certvit} utilize Lipschitz continuous $L_2$-attention ViT architectures which are pre-trained on ImageNet-1K and use them for certification on CIFAR10/100~\citep{cifar10}, MNIST~\citep{deng2012mnist} and Tiny ImageNet~\citep{Le2015TinyIV} respectively. 
Their proposed algorithm involves a proximal step that initially constrains the weights thereby reducing the model Lipschitz at each layer followed by a projection step that focuses to improve the generalization, hence the clean accuracy of the model. They use the power method to constrain the Lipschitz constant of the successive layers. To the best of our knowledge, this is the only work that has reported the model Lipschitz of a vision transformer-based architecture.
As part of their main findings, the authors explain that CertViT significantly reduces the Lipschitz bounds of networks compared to existing Lipschitz methods like GloRo, Local-Lip, and BCP. Their results on  MNIST, CIFAR-10, CIFAR-100, and TinyImageNet~\citep[Table 1]{gupta2023certvit} shows that CerViT achieves superior performance for both clean and adversarial training (trained on PGD~\citep{madry2018towards} samples). 
Although their technique proves to be better than existing Lipschitz based certification methods, their two-step algorithm requires several iterations to converge and does not seem to scale with increasing dataset complexity. 

LipsFormer~\citep{qi2023lipsformerintroducinglipschitzcontinuity} introduces a new vision transformer architecture that is Lipschitz continuous. 
They update the existing 
SwinV2~\citep{liu2022swintransformerv2scaling} architecture to be Lipschitz continuous by replacing core elements such as the non-overlapping window attention mechanism with the Scaled Cosine Similarity Attention (SCSA)~\citep{qi2023lipsformerintroducinglipschitzcontinuity} and replaing the LayerNormat each stage with their proposed CenterNorm~\citep{qi2023lipsformerintroducinglipschitzcontinuity} layer which they provide theoretical guarantees for to be 1-Lipschitz normalization layer. 
Furthermore, they also explain how DropPath~\citep{goyal2017accurate} in addition to DropOut helps in tightening the Lipschitz upper bound. 
Although, they provide experimental results on the performance of Lipschitz continuous vision transformer on ImageNet-1K, they do not provide an emperical estimation of the model Lipschitz  or robust accuracy measure that can be used as a benchmark.

Considering the drawbacks of the works presented in this section, in the next section we propose our method that tries to overcome some of the gaps highlighted in the existing research in this field.

\section{Methodology}\label{sec:method}

To obtain a tight upper bound on the model Lipschitz, we introduce some changes to the original ShiftViT~\citep{wang2022shiftoperationmeetsvision} architecture. 
This mean according to~\autoref{eq: lipschitz-condition}, we replace architectural components that have a higher $L$ comparatively to obtain a model Lipschitz as low as possible. A low model Lipschitz ensures that the model output is deterministically bounded for any input $\mathbf{x} \in \mathbb{R}^{N\times N}$ and is easier to certify using norm-based verification.

\subsection{Lipschitz Continuous Modules}

\paragraph{CenterNorm instead of GroupNorm}

GroupNorm~\citep{wu2018group}, defined in~\autoref{eq: groupnorm} plays an important role in the ShiftViT~\citep{wang2022shiftoperationmeetsvision} architecture to provide stable training and enhanced generalization performance. Its effectiveness is in handling batches of small sizes by computing batch statistics across channel groups, defined as: 
\begin{equation} \label{eq: groupnorm}
    GN(\mathbf{x}) = \gamma \left( \frac{\mathbf{x} - \mu}{\sqrt{\sigma^2 + \epsilon}} \right) + \beta,
\end{equation}

where $\mu$ and $\sigma^2$ are the mean and variance computed across the channels within each group, $\gamma$ and $\beta$ are learnable parameters, and $\eta$ is a small constant for numerical stability. In a Lipschitz constrained training setting, GroupNorm operation becomes difficult to optimize as the denominator term, $\eta$ is of the order $10^{-6}$. As training progresses, we see that since $\eta$ is inversely proportional to the Jacobian of $GN(\mathbf{x})$, it proves to be a bottleneck for stable training under Lipschitz constraint.

CenterNorm~\citep{qi2023lipsformerintroducinglipschitzcontinuity}, proposed in the LipsFormer~\citep{qi2023lipsformerintroducinglipschitzcontinuity} architecture, is a 1-Lipschitz normalization that works as a viable alternative to GroupNorm. CenterNorm is mathematically formulated as:
\begin{equation}
    CN(\mathbf{x}) = \gamma \cdot \left( \textbf{I} - \frac{1}{D} \mathbf{1}\mathbf{1}^T \right) \mathbf{x} + \beta
\end{equation}

where $D$ is the input dimension, $\textbf{I}$ is the identity matrix. $\gamma$ and $\beta$ are learnable parameters initialized as $1$ and $0$ respectively. $\mathbf{1}\mathbf{1}^T$ is a matrix of ones. This formulation is Lipschitz continuous as shown by~\citet{qi2023lipsformerintroducinglipschitzcontinuity} with a theoretical Lipschitz upper bound of $\frac{D}{D-1}$. For practical purposes, the Lipschitz constant can be approximated to 1 since he input dimension $D$ is high for datasets like CIFAR10~\citep{cifar10}. In our experiments, we compute the Lipschitz constant of CenterNorm as the as the absolute maximum of the input weight matrix as we want a tight bound. 

Unlike GroupNorm~\citep{wu2018group}, CenterNorm~\citep{qi2023lipsformerintroducinglipschitzcontinuity} directly eliminates the mean across dimensions but does not scale by the standard deviation, thereby preventing instability from small variance values. It ensures a limited gradient, offering a more stable and robust training experience, which accelerates convergence and enhances generalization in LipShiFT. The combination of these theoretical benefits and empirical evidence of smoother training trajectories without the necessity for learning rate warmup establishes CenterNorm as a preferred normalization method in vision transformer architectures compared to GroupNorm.

\paragraph{MaxMin instead of GeLU}
To empirically reduce the upper bound of model Lipschitz, we replace the GeLU activation which has a Lipschitz bound of approximately 1.12~\citep{qi2023understanding} with a 1-Lipschitz MinMax activation function. 
This has essentially proven to be extremely useful in improving model robustness in past works~\citep{hu2023scaling, anil2019sorting}. As \citet{anil2019sorting} explains, MinMax displays gradent preserving property during adversarial training that is valuable for training models to be more robust and generalize better.

\paragraph{Linear Residual Convolution (LiResConv) instead of MLP}

\begin{figure}[htbp]
\centering
\includegraphics[width=0.6\textwidth, trim = 8cm 6cm 9cm 3.5cm, clip]{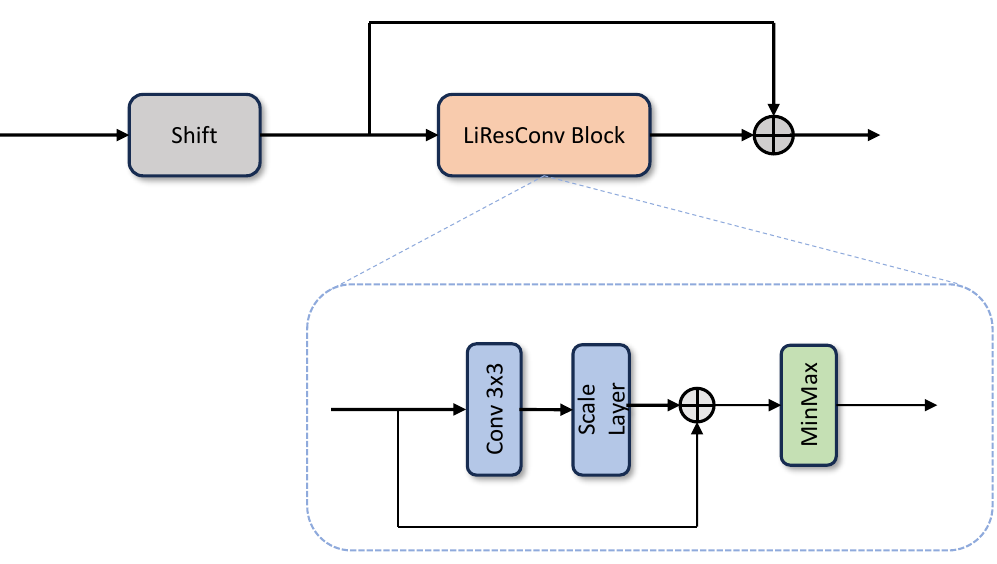}
\caption{Proposed modified shift block.}
\label{fig:architecture_new}
\end{figure}

Another core architectural component of the ShiftViT~\citep{wang2022shiftoperationmeetsvision} is a Multi Layer Perception (MLP) within the shift block. It consists of three alternating 1x1 convolutional and GeLU activation layers followed by a DropOut layer. This MLP network is Lipschitz continuous since it is a composition of affine transformations with activation layers in between~\citep{pauli2023lipschitz}. 

But as \citet{hu2023scaling} motivates, the LiResConv block proves to have a tighter upper bound compared to standard FFN \citep[Section 3]{hu2023scaling}.

\paragraph{AveragePool instead of AdaptiveAveragePool}

Since the computation of the model Lipschitz constant is dependent on the input size, we replace the final pooling layer AdaptiveAveragePool2d in the ShiftViT~\citep{wang2022shiftoperationmeetsvision} architecture with the standard AveragePool2d layer. We compute the Lipschitz constant using the power iteration method~\citep{farnia2018generalizableadversarialtrainingspectral}. We use the custom  implementation of AveragePool2d with Lipschitz computation by \citet{hu2023scaling}.  

\paragraph{Dropout and DropPath} \label{paragaph: drop-effect}

\citet{gouk2021regularisation} studies the effect of using regularization techniques in scalable DL architectures. From a Lipschitz perspective, the effect of DropOut and DropPath are important to take into consideration.
Both regularization techniques play an essential role in regularizing ShiftViT~\citep{wang2022shiftoperationmeetsvision}. 
For Lipschitz training of LipShiFT, we scale the model Lipschitz during inference by a factor of $(1-P_{drop})$ where $P_{drop}$ is the drop out rate used throughout the model. This way we can empirically compensate for the regularization applied during training with EMMA~\citep{hu2023scaling} loss and ensure a tight upper bound on the overall model Lipschitz.

\paragraph{Orthogonal Initialization for Affine Layers}

Initialization plays an important role in successfully training a neural network. Many initialization methods have been proposed in the past years such as \citet{glorot2010understanding} and \citet{he2015delving} initialization. 
We use orthogonal initialization which is 1-Lipschitz by design since the initialized weight matrix is orthogonal,

\begin{equation}
\mathbf{W}_{orth} = \frac{\mathbf{W}_{init}}{\sigma_{max}(\mathbf{W}_{init})} 
\end{equation}

where $\mathbf{W}_{orth}$ is the initial random orthogonal weight matrix and $\sigma_{max}(\mathbf{W}_{init})$ is its largest eigenvalue. For affine transformation $\textbf{f} (\mathbf{x}, \mathbf{W}_{orth}) = \mathbf{W}_{orth}^{T}\mathbf{x}$, its Lipschitz constant satisfies the following inequality, $||\mathbf{W}_{si}^{T}\mathbf{x}_{1} - \mathbf{W}_{orth}^{T}\mathbf{x}_{2}|| \leq Lip(\textbf{f}_{\mathbf{x}}(\mathbf{W}_{orth}))||\mathbf{x}_1 - \mathbf{x}_2||$, where $Lip(\textbf{f}_{\mathbf{x}}(\mathbf{W}_{orth})) = 1$ at initialization. We use orthogonal initialization on all affine layers.

\paragraph{Last Layer Normalization (LLN)}

Lastly, we replace the standard linear prediction head with a normalized prediction head introduced by \citet{singla2021improved}. 
They state that it particularly is useful when the input dataset has a large number of class labels, hence helps to achieve robust certificates at scale. Further, this has been seen to improve emperical robustness of the ResNet based LiResNet~\citep{hu2023scaling} architecture.

\section{Experimental Results}\label{sec:experimental_results}

This section describes the main results of LipShiFT under adversarial constraints. We briefy describe the training and evaluation settings and discuss our main findings. Lastly, we report robust accuracy of our LipShift model using the AutoAttack~\citep{croce2020reliable} framework. 
We report our settings in~\autoref{app:appendix}.

\subsection{Certified Robustness} \label{sec:eval}

\begin{table*}[htb]
  \centering
  \small
  \caption{This table presents the clean and verified robust accuracy (VRA) of several concurrent works and our LipShiFT on CIFAR-10/100 and Tiny ImageNet datasets.}
  \label{table: main_results}
    \adjustbox{max width=\textwidth}{%
    \begin{tabular}{l|l|ccr|rr}
        \toprule
        \textbf{Dataset} & \textbf{Method}   & \textbf{Architecture} & \textbf{Specification} & \textbf{\#Param.} & \textbf{Clean} & \textbf{VRA} \\
        & & & &(M) &(\%) &(\%) \\
        \midrule
        \multirow{9}{*}{\textbf{CIFAR-10}} & BCOP~\scriptsize{\citep{li2019preventinggradientattenuationlipschitz}}               & ConvNet      & 6C2F  & 2 &  75.1    &  58.3 \\
        & GloRo~\scriptsize{\citep{leino21gloro}}           & ConvNet      & 6C2F  & 2 & 77.0    & 60.0     \\
        & Local-Lip Net~\scriptsize{\citep{huang21local_lipschitz}}         & ConvNet      & 6C2F  & 2 & 77.4    & 60.7     \\
        & Cayley~\scriptsize{\citep{trockman21orthogonalizing}}            & ConvNet      & 4C3F   & 3 & 75.3    & 59.2     \\
        & SOC (HH+CR)~\scriptsize{\citep{soc}}               & ConvNet      & LipConv-20  & 27 & 76.4    & 63.0     \\
        & CPL~\scriptsize{\citep{CPL}}            & ResNet      & XL &  236 & 78.5    & 64.4     \\
        & SLL~\scriptsize{\citep{araujo2023a}}            & ResNet      & XL  & 236 & 73.3    & 65.8     \\
        & LiResNet (+DDPM) ~\scriptsize{\citep{hu2023scaling}}& ResNet      & 12L512W & 49 & 82.1 & 70.0 \\
        & CHORD LiResNet (+DDPM) ~\scriptsize{\citep{hu2024a}}& ResNet      & 12L512W & 49  & \textbf{87.0} & \textbf{78.1}  \\
        & CertViT ~\scriptsize{\citep{gupta2023certvit}} & ViT & 6-layer & 41  & 75.1 & 33.1 \\
        & LipShiFT (ours)&  ViT      & [6,6,10,6] & 49  & 71.8 & 63.2  \\
        & LipShiFT (+DDPM) (ours)&  ViT      & [6,6,10,6] & 49  & 73.6 & 65.2  \\
        \midrule
        \midrule
        \multirow{7}{*}{\textbf{CIFAR-100}} & BCOP~\scriptsize{\citep{li2019preventinggradientattenuationlipschitz}}                & ConvNet      & 6C2F & 2 &  45.4    &  31.7 \\
        & Cayley~\scriptsize{\citep{trockman21orthogonalizing}}               & ConvNet      &  4C3F  & 3 & 45.8   &   31.9  \\
        & SOC (HH+CR)~\scriptsize{\citep{soc}}               & ConvNet      & LipConv-20  & 27 & 47.8   & 34.8     \\
        & LOT~\scriptsize{\citep{xu2023lotlayerwiseorthogonaltraining}}               & ConvNet      & LipConv-20  & 27 & 49.2   & 35.5     \\
        & SLL~\scriptsize{\citep{araujo2023a}}               & ResNet      & XL  & 236 & 46.5   & 36.5     \\
        & LiResNet (+DDPM)  ~\scriptsize{\citep{hu2023scaling}}& ResNet     & 12L512W & 49& 55.5 & 41.5  \\
       & CHORD LiResNet (+DDPM) ~\scriptsize{\citep{hu2024a}}& ResNet     & 12L512W & 49& \textbf{62.1} & \textbf{50.1} \\
       & CertViT ~\scriptsize{\citep{gupta2023certvit}} & ViT & 10-layer & 49  & 46.2 & 9.1 \\
       & LipShiFT (ours)& ViT     & [6,6,10,6] & 49  & 43.3 & 34.1  \\
        & LipShiFT (+DDPM)(ours)& ViT      & [6,6,10,6] & 49  & 44.5 &  35.8 \\
        \midrule 
        \midrule
        \multirow{3}{*}{\textbf{Tiny ImageNet}} & LiResNet(+DDPM)~\scriptsize{\citep{hu2023scaling}} & ResNet  & 12L512W & 49  & 46.7    &  33.6        \\
        & CHORD LiResNet (+DDPM) ~\scriptsize{\citep{hu2024a}}& ResNet     & 12L512W & 49& \textbf{48.4} & \textbf{37.0} \\
        & CertViT ~\scriptsize{\citep{gupta2023certvit}} & ViT & 12-layer & 91  & 36.3 & 2.3 \\
       & LipShiFT(ours)& ViT      & [6,6,10,6] & 49  & 36.2 &  28.1 \\
        \bottomrule
    \end{tabular}
    } 
\end{table*}%

In~\autoref{table: main_results}, we compare the main results of our proposed work in to existing techniques and methodologies implemented in the recent years. For a fair comparison, we report our main results on a custom LipShiFT specification that has the same number of parameters when comparing to works such as the CHORD-LiResNet~\citep{hu2024a}. 
We compare performance of our architecture with recent works on the CIFAR10/100~\citep{cifar10} and Tiny ImageNet~\citep{Le2015TinyIV} datasets respectively. 
For each of the datasets, we report the clean and verified accuracy. The verified accuracy (VRA) is reported for a target epsilon of $\epsilon = 36/255$. The best scores are marked in bold.

In~\autoref{table: main_results}, except CertViT~\citep{gupta2023certvit}, LipShiFT is the only architecture that utilizes a  transformer-based architecture and uses Lipschitz constrained techniques to provide robustness guarantees. We also stand out from previous works as we are able to report a tight upper bound of the empirical model Lipschitz constant. As has been successfully shown in case of LiResNet~\citep{hu2023scaling, hu2024a}, we implement a similar Lipschitz constrained training scheme since it serves to be an efficient way to constrain the model Lipschitz while being able to provide deterministic and certifiable robustness scores. 

We also report our results alongside previous works that use different architectures for each dataset. These are broadly divided into three categories, namely, ConvNets, ResNets and ViT respectively. For each architectural type, we also provide specification detail of the models that yielded the corresponding results. In ConvNets, $C$ represents the number of convolution layers, and $F$ indicates the number of fully connected layers. In ResNets, we employ the conventional notation where $L$ denotes the number of layers, and $W$ signifies the width of the dense layer. For ViT based architectures, we add results from CertViT~\citep{gupta2023certvit} which uses an $L_2$-attention transformer that has a depth of 6-layers for CIFAR10 and 12-layers for CIFAR100 respectively. LipShiFT is based on the SwinV2~\citep{liu2022swintransformerv2scaling} transformer which introduces network depth split into repetitive stages which changes the LiShiFT specification compared to the one used for CertViT~\citep{gupta2023certvit}. So for example, $[6, 6, 10, 6]$ defines the depth of the shift block at each stage.

To incorporate augmented images for training, we follow \citet{hu2024a} and use a $1:3$ ratio for clean to augmented images per batch. This applies to both CIFAR10 and CIFAR100 datasets. As already mentioned, we do not use augmented data for training LipShiFT on Tiny ImageNet. As noted by~\cite{hu2024a} for a ResNet architecture, we find the effect of adversarially training a transformer-based model with additional augmented data is overall positive and see increments across datasets and accuracy metrics. On CIFAR10, we see a $2.08\%$ increase in verified accuracy and $1.03\%$ increase in clean accuracy for the same  model specification. Similarly on CIFAR100, we see a increase on clean accuracy and verified accuracy by approximately $1.20\%$ and $1.62\%$ respectively.  

Comparing our results to CertViT~\citep{gupta2023certvit}, we find that LipShiFT comprehensively outperforms on all three datasets on verified accuracy. On Tiny ImageNet, LipShiFT is able to outperform with a model size that is nearly 2x smaller. Compared to ConvNet based approaches, our LipShiFT model (non DDPM version) outperforms works such as~\citet{li2019preventinggradientattenuationlipschitz, leino21gloro, huang21local_lipschitz, trockman2021orthogonalizingconvolutionallayerscayley} and~\citet{soc} respectively on verified accuracy while not on clean accuracy. Moreover, our LipShiFT currently does not outperform popular ResNet approaches such as LiResNet~\citep{hu2023scaling, hu2024a} while being comparable to works such as \citet{meunier2022dynamicalperspectivelipschitzneural} and \citet{araujo2023a} despite being nearly 5x smaller. 
This comparison shows us that LipShiFT performs competitively not just compared to transformer-based architectures but when also when compared to ConvNets and ResNets.

\begin{figure}[htb] 
    \centering
     \begin{subfigure}[b]{0.45\textwidth}
    \centering
        \includegraphics[width=\textwidth]{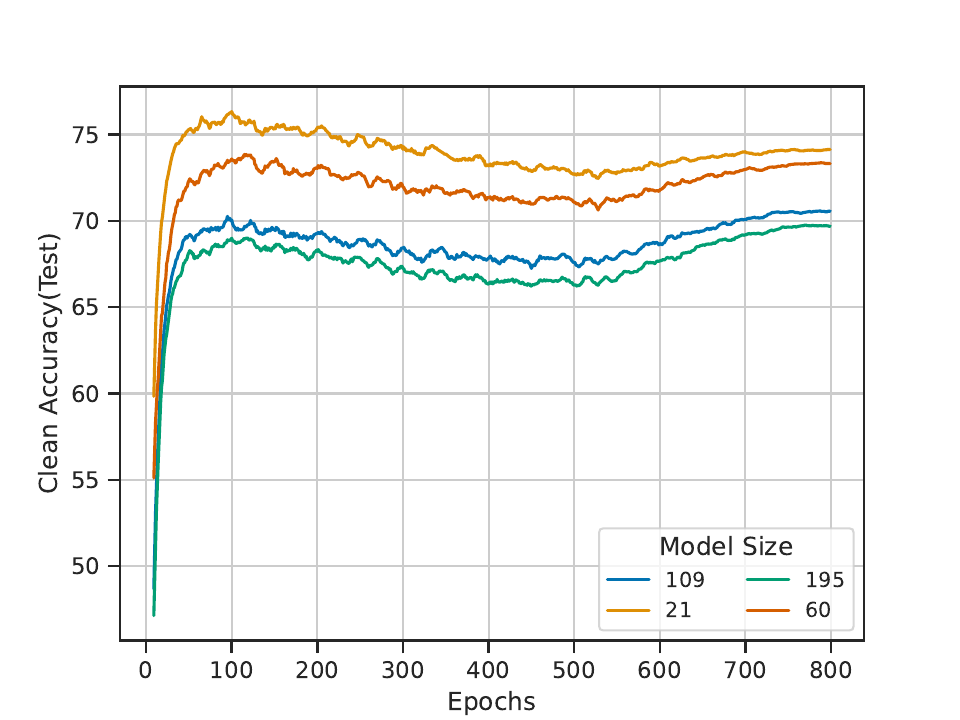}
        \caption{Clean accuracy.}    
        \label{fig:ablation-ms-accuracy}
    \end{subfigure}
    \begin{subfigure}[b]{0.45\textwidth}
    \centering
        \includegraphics[width=\textwidth]{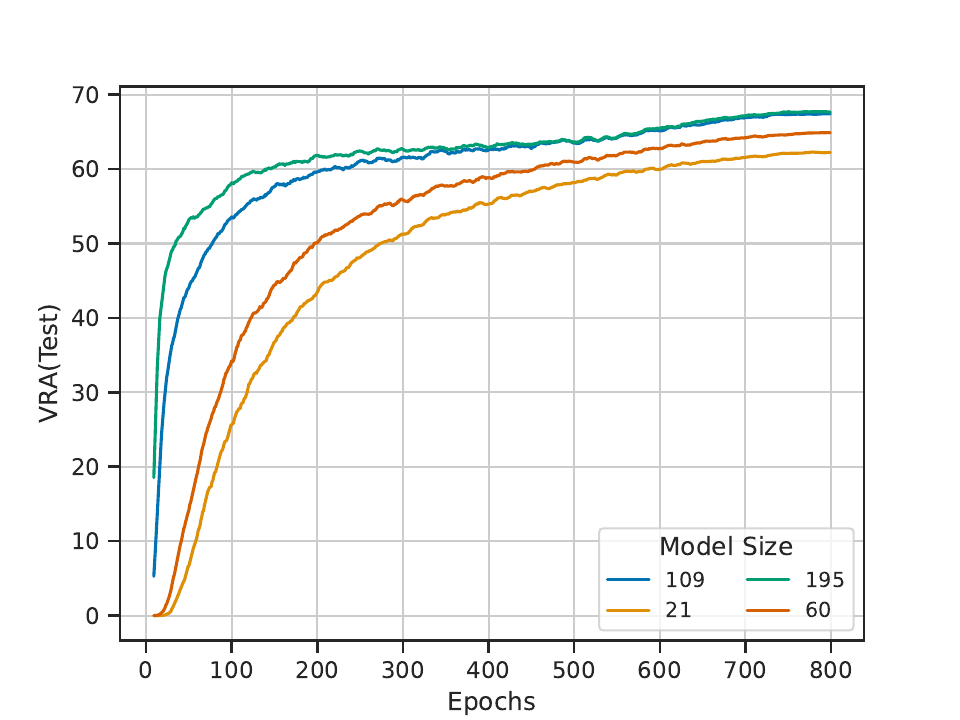}
    \caption{Verified accuracy.}
    \label{fig:ablation-ms-vra}
    \end{subfigure}
    \caption{Effect of model size on verified and clean accuracy.}
    \label{fig: model-size-overall}
\end{figure}

In~\autoref{fig: model-size-overall}, we sample four model sizes where the legend depicts the number of trainable parameters (in millions) for each LipShiFT model configuration. The smallest LipShiFT model we test with ShiftViT-T(Light) (21M) and the largest ShiftViT-B (195M). The number of parameters will vary for corresponding LipShiFT and ShiftViT model sizes due to the various architectural changes we introduce in LipShiFT. For verified accuracy in~\autoref{fig:ablation-ms-vra}, we see that as the model size increases, it becomes more robust. For this ablation, we also train the models for 800 epochs, which is more than all other experiments. This is because we find that LipShiFT-S (109M) and LipShiFT-B (195M) do not converge beyond even after 800 epochs while LipShiFT-T(light) (21M) and LipShiFT-T (60M) converge around the $500^{th}$ epoch. This also shows that our approach is scalable to larger transformer-based networks.

\begin{table}[htb] 
  \centering
  \caption{This table presents the Lipschitz constants for our LipShiFT models reported in the~\autoref{table: main_results}.}
  \label{table:main_results_model_lipschitz}
    \begin{tabular}{llcc}
    \toprule
        \textbf{Dataset} & \textbf{Method} & \textbf{\makecell[c]{Lip. Constant}} \\
        \midrule
        \multirow{2}{*}{\textbf{CIFAR-10}} 
        & LipShiFT  & 192.0  \\
        &LipShiFT (+DDPM)  & 427.2  \\
        \midrule
        \multirow{2}{*}{\textbf{CIFAR-100}} 
       & LipShiFT & 47.7  \\
        & LipShiFT (+DDPM) &  68.61 \\
        \midrule
        \multirow{1}{*}{\textbf{Tiny ImageNet}}
       & LipShiFT & 26.69  \\
        \bottomrule
    \end{tabular}
\end{table}%

We additionally report the Lipschitz constants in~\autoref{table:main_results_model_lipschitz}. These are corresponding to the model configurations reported in~\autoref{table: main_results}. Based on that, we can see that the Lipschitz constants generally decrease as the complexity of the dataset increases. While for the models trained on CIFAR datasets, we see the LipShiFT (+DDPM) has higher model Lipschitz compared to one trained without augmented data.  
Additional ablation studies are reported in \autoref{app: apendix_ablations}.

\subsection{Empirical Robustness}

\begin{table*}[t]
\centering
  \caption{This table presents the clean and empirical robust accuracy of several concurrent works and our LipShiFT model on CIFAR-10/100 and Tiny ImageNet datasets respectively.}
    \adjustbox{max width=\textwidth}{%
    \begin{tabular}{llcc}
    \toprule
     \multicolumn{1}{c}{\multirow{2}[2]{*}{\textbf{Datasets}}}  & \multicolumn{1}{c}{\multirow{2}[2]{*}{\textbf{Models}}} & \multicolumn{1}{c}{\multirow{2}[2]{*}{\textbf{Clean} \textbf{Accuracy}}} & \multicolumn{1}{c}{\multirow{2}[2]{*}{\textbf{AutoAttack} $\epsilon=36/255$}} \\
     \\
    \midrule
     \multirow{5}[4]{*}{\textbf{CIFAR-10}}
      & CPL XL~\scriptsize{\cite{CPL}} & 78.5 & 71.3  \\
      & SLL X-Large~\scriptsize{\cite{araujo2023a}}& 73.3 & 70.3  \\
       & GloRo CHORD LiResNet~\scriptsize{\cite{hu2024a}} & \textbf{87.0} & \textbf{82.3}  \\
        & LipShiFT (Ours)& 71.8 & 65.0  \\
        & LipShiFT(+DDPM) (Ours)& 73.7 & 66.5  \\
      
    \midrule
    \multirow{6}[4]{*}{\textbf{CIFAR-100}}

      & CPL XL~\scriptsize{\cite{CPL}} & 47.8 & 39.1  \\
      & SLL Large~\scriptsize{\cite{araujo2023a}}& 54.8 & 44.0 \\
      & Sandwich~\scriptsize{\cite{wang2023direct}} & 57.5 & 48.5 \\
       & GloRo CHORD LiResNet~\scriptsize{\cite{hu2024a}}& \textbf{62.1} & \textbf{55.5} \\
        & LipShiFT (Ours)& 43.3 & 37.0 \\
       & LipShiFT(+DDPM) (Ours)& 44.5 & 37.8 \\
    \midrule
    \multirow{3}[3]{*}{\textbf{Tiny ImageNet}}
    
    & SLL Medium~\scriptsize{\cite{araujo2023a}} & 30.3 & 24.6 \\
      & Sandwich Medium~\scriptsize{\cite{wang2023direct}} & 35.5 & 29.9 \\
       & GloRo CHORD LiResNet (+DDPM)~\scriptsize{\cite{hu2024a}}& \textbf{48.4} & \textbf{42.9} \\
    & LipShiFT (Ours)& 36.2 & 32.1\\
    \bottomrule
    \end{tabular}
    }
    \label{table:autoattack-results}
\end{table*}%

We compute empirical robustness using the AutoAttack~\citep{croce2020reliable} framework and summarize our results in~\autoref{table:autoattack-results}. The framework allows an adversarial attack ensemble for robustness assessment on four types of attacks. These include two step-size free variants of PGD attacks known as APGD-CE and APGD-DLR. They are variants of Auto-PGD attack what use Cross Entropy and DLR loss respectively. Also including FAB-T attack that minimizes the adversarial perturbation norm and Square attack that is a query efficient black-box attack. As we report our results along with notable works from the past, we see that very few works have reported AutoAttack~\citep{croce2020reliable} on Tiny ImageNet. 

We find that the LipShiFT performance under the framework attack ensemble provides competitive results when compared to ConvNet and ResNet architectures. Specifically, compared to CPL~\citep{CPL} and SLL X-Large~\citep{araujo2023a} models on CIFAR10, are  better by 6.26\% and 5.26\%  respectively compared to our non-DDPM trained LipShiFT model. On the CIFAR100 dataset, we see that this gap in robust accuracy is at  2.1\% and 7.0\% respectively. On Tiny ImageNet, we see that LipShiFT outperforms both ResNet based SLL Medium~\citep{araujo2023a} and Sandwich Medium~\citep{wang2023direct} models by 7.48\% and 2.18\% respectively. Currently, our LipShiFT models aren't competitive yet with the state-of-the-art CHORD LiResNet models across any of the datasets. Nevertheless, just like the result comparison in~\autoref{table: main_results}, we find that the LipShiFT performance on AutoAttack~\citep{croce2020reliable} can be competitive to current state-of-the-art models when measuring empirical robustness using AutoAttack~\citep{croce2020reliable}.

\section{Conclusion}\label{sec:conclusion}

 Vision transformers are extremely versatile as proven by past research for various applications and has been proven time and again to be useful to process text and image based data efficiently. While they have been used excessively in various application domains, a study from a formal verification perspective is critically lacking as compared to architectures such as ConvNets and ResNets.    

The work presented in this paper focuses on one particular branch of formal verification know as Lipschitz based certification and applies it to vision transformer-based architecture to provide a tight and empirically sound assessment of the Lipschitz constant of the model. We surpass well known bottlenecks in transformer robustness certification by utilizing a shift-based alternative to the self-attention mechanism that is parameter free and also 1-Lipschitz while maintaining generalization performance. Then we implement a series of architectural changes to  ShiftViT~\citep{wang2022shiftoperationmeetsvision} and propose a new, more Lipschitz focused vision transformer known as LipShiFT. To constrain the Lipschitz constant of this model, we perform adversarial training using EMMA loss introduced by \citet{hu2023scaling}.

Using our method, we are able to successfully develop a robust vision transformer that is lightweight and scalable over different datasets. We also show that our method provides sound results for larger model sizes and provide robustness certification for them too. In addition, we also see the effects of training our LipShiFT model using additional augmented data. Finally, we verify our finding by reporting certified robustness scores of LipShiFT in~\autoref{table:autoattack-results} using the AutoAttack framework across all three datasets. This sets up a promising foundation for more focus on certifying vision transformers and bridge the gap in research that currently exists in this sub-branch of formal verification.

\bibliography{references}
\bibliographystyle{iclr2025_conference}

\appendix
\section{Appendix}\label{app:appendix}

\subsection{Datasets} 

We report our main results on the CIFAR10/100~\citep{cifar10} and Tiny ImageNet~\citep{Le2015TinyIV}. 
For the CIFAR datasets, we use the standard input resolution of 32 while for Tiny Imagenet, we use 64 respectively. For all our experiments, we apply RandomCrop augmentation with a padding of 4. We additionally make use of augmented data generated in the work of  \citet{wang2023betterdiffusionmodelsimprove}  using the image generation process defined in Elucidating Diffusion Models (EDM)~\citep{karras2022elucidatingdesignspacediffusionbased}. To achieve this, \citet{wang2023betterdiffusionmodelsimprove} utilized 4 NVIDIA A100 SXM4 40GB GPUs for training and image generation. The training scripts use a 64-bit version of Python 3.8 and PyTorch 1.12.0. For CIFAR-10, they generate images using the pre-trained model from EDM~\citep{karras2022elucidatingdesignspacediffusionbased}, which yields a new state-of-the-art FID of 1.79. For 5M data generation, following \citet{rebuffi2021fixingdataaugmentationimprove}, they generate 500K images for each class.
For CIFAR100, \citet{wang2023betterdiffusionmodelsimprove} obtain the best FID (2.09) after 25 sampling steps by training their own model on the same 4 NVIDIA A100 cluster. For CIFAR100, they generate 50M samples out of which we select 5M in total using random indexing. All classes are sampled equally.
For each of the CIFAR datasets, we utilize 5M generated samples for training LipShiFT versions differentiated by (+DDPM). For training LipShiFT on Tiny ImageNet, we do not use any additional generated data.

\subsection{Hardware details} 

Each experiment is run a single machine with NVIDIA RTX 4090 GPU with 24GB of VRAM for each of the CIFAR and Tiny ImageNet datasets. Our implementation is based on PyTorch~\citep{paszke2019pytorchimperativestylehighperformance}.

\subsection{Training Settings}

For both the CIFAR datasets and Tiny ImageNet, the models are trained with the AdamW ~\citep{loshchilov2019decoupledweightdecayregularization} with a batch size of 128 and a learning rate of $5e^{-4}$ for 500 epochs. We use a cosine learning rate decay ~\citep{loshchilov2017sgdrstochasticgradientdescent} with no warmup. Results reported with (+DDPM) denote that the model is trained with augmented data, none otherwise.

\subsection{Epsilon Schedule}

As introduced by \citet{hu2023scaling} in their work, we utilize a similar variable $\epsilon$ strategy with EMMA loss for training LipShiFT. 

Let the total number of epochs is $T$ and the certification radius be $\epsilon$, varies as follows,
\begin{equation} \label{eq: variable-eps}
\epsilon_{\text{train}}(t) = \left(\min(\frac{3t}{2T}, 1)\times 1.0 \right) \epsilon, \quad \epsilon=36/255.
\end{equation}
at epoch $t \in [0, T]$. As a result, $\epsilon_{\text{train}}(t)$ begins at $0.0$ and increases linearly to $\epsilon$ before arriving three quarters of the way through the training. Later, $\epsilon_{\text{train}}$ remains $\epsilon$ to the end. 
We do not try to overshoot this range since our target certification radius is $\epsilon$. 
We use the same schedule across all our experiments. 
The variable $\epsilon$ schedule is used only during training along with EMMA loss as at test time, we do not need to push the decision boundary for certification. 
During evaluation, we use the target $\epsilon=36/255$ along with TRADES loss. 
As described in~\citet{leino21gloro}, we use the default $\lambda = 1$ for all our experiments.

\section{Additional Ablation Studies}{\label{app: apendix_ablations}}

To supplement the model size ablation depicted in~\autoref{fig: model-size-overall}, we conducted additional experiments to observe the effect of some standard hyperparameters on the accuracy metrics for LipShiFT. The following studies are reported for a LipShiFT-T(light) (21M) model. Unlike the model size ablation, we run the model training for 500 epochs.

\begin{figure}[htb]
    \centering
    \begin{subfigure}[b]{0.45\textwidth}
        \centering
        \includegraphics[width=\textwidth]{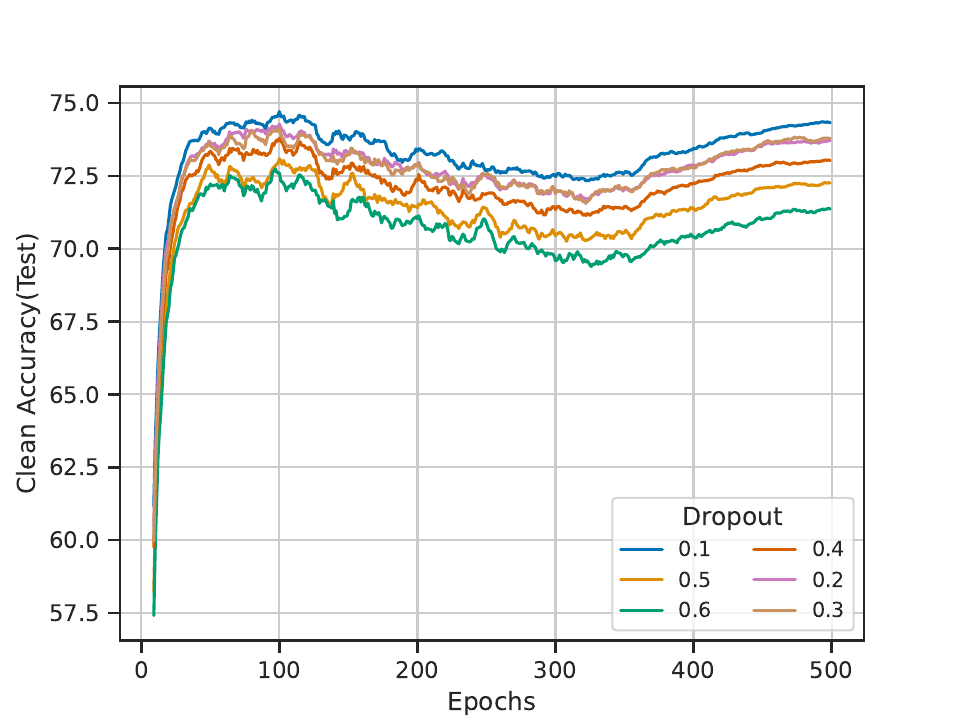}
        \caption{Clean Accuracy}
        \label{fig:ablation-drop-clval}
    \end{subfigure}
    \hfill
    \begin{subfigure}[b]{0.45\textwidth}
        \centering
        \includegraphics[width=\textwidth]{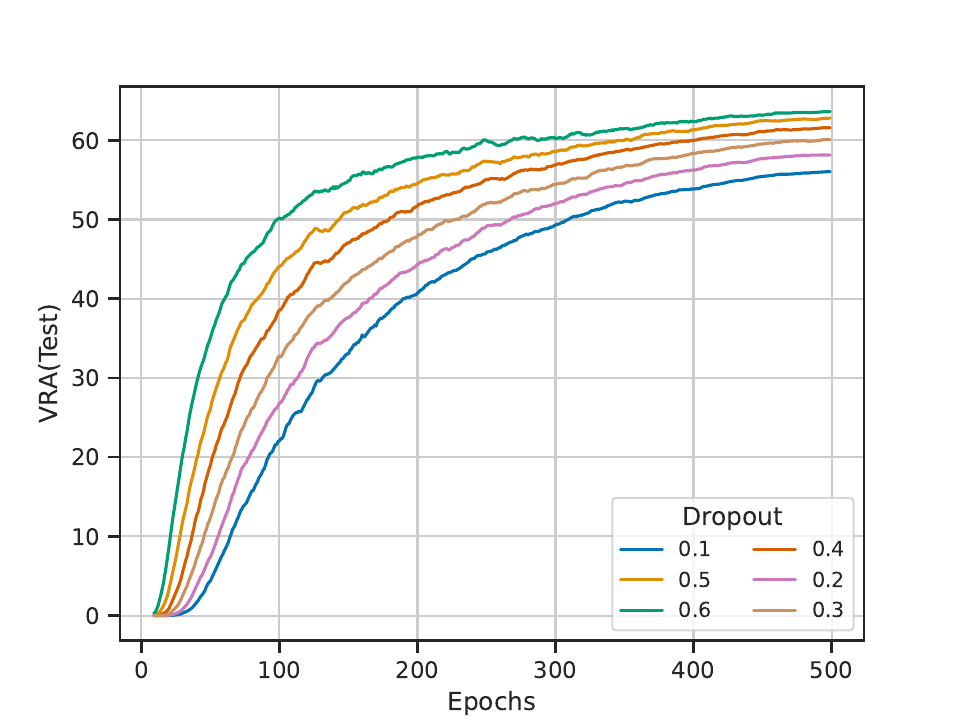}
        \caption{Verified Accuracy}
        \label{fig:ablation-drop-vraval}
    \end{subfigure}

    \caption{Effect of dropout rate on accuracy.}
    \label{fig:ablation-drop}
\end{figure}

\paragraph{Effect of Dropout Rate}
We perform experiments using dropout ranging from $[0.1,0.6]$ with a step increment of $0.1$.~\autoref{fig:ablation-drop-clval} shows the clean accuracy for different rates of dropout. We observe that $p_{drop} = 0.6$ is a very high regularization for Lipschitz margin training for LipShiFT model as the clean accuracy is the lowest throughout the training compared to other droput rates. Although, interestingly, the verified accuracy for $p_{drop} = 0.6$ is highest and decreases along with $p_{drop}$. This essentially is due to the dropout normalization we perform while computing the model Lipschitz. Overall,~\autoref{fig:ablation-drop} depicts the intrinsic difficulty of the model learning to improve the clean and verified accuracy simultaneously.

\begin{figure}[htb]
    \centering
    \begin{subfigure}[b]{0.45\textwidth}
        \centering
        \includegraphics[width=\textwidth]{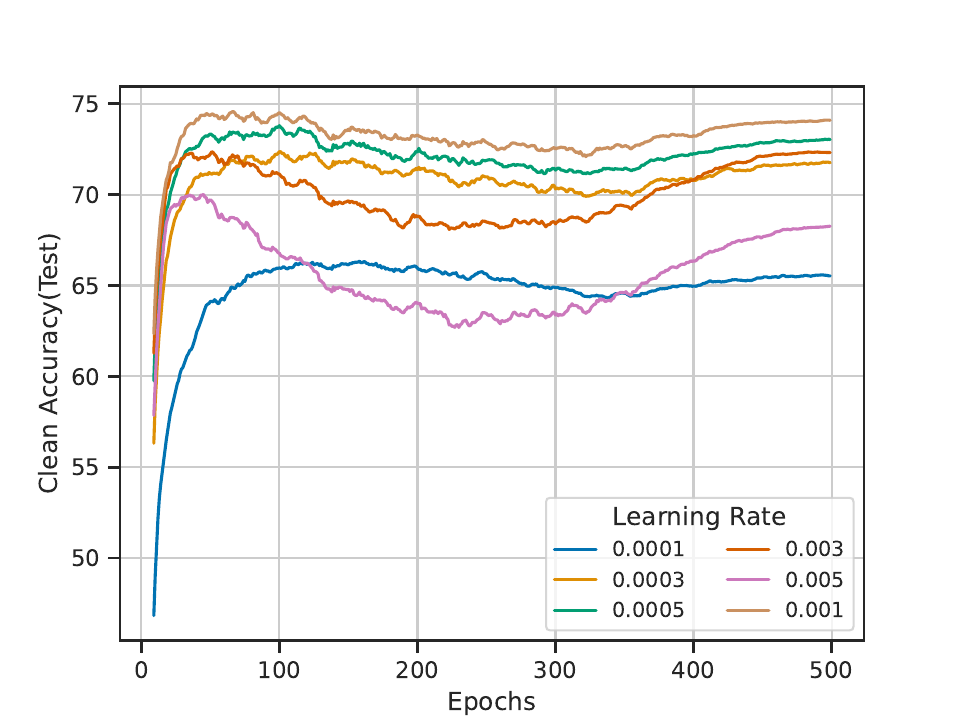}
        \caption{Clean Accuracy}
        \label{fig:ablation-lr-clal}
    \end{subfigure}
    \hfill
    \begin{subfigure}[b]{0.45\textwidth}
        \centering
        \includegraphics[width=\textwidth]{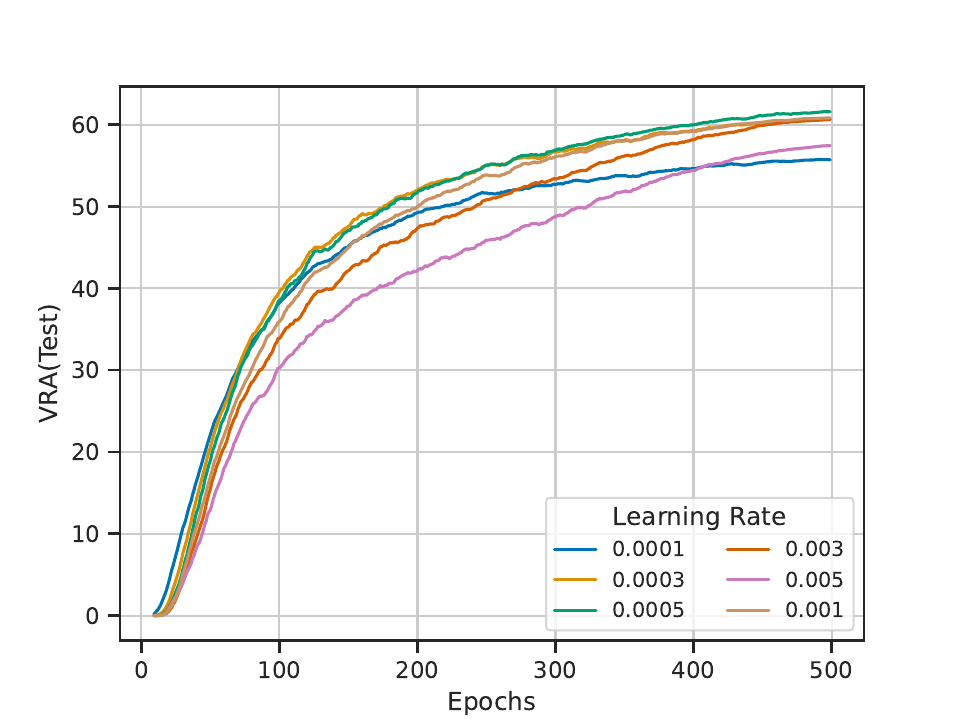}
        \caption{Verified Accuracy}
        \label{fig:ablation-lr-vraval}
    \end{subfigure}

    \caption{Effect of learning rate on accuracy.}
    \label{fig:ablation-lr}
\end{figure}

\paragraph{Effect of Learning Rate}
We perform experiments using learning rate in this fixed range [0.001, 0.003, 0.005, 0.0001, 0.0003, 0.0005].~\autoref{fig:ablation-lr} shows the overall effect of learning rate on clean accuracy and VRA for LipShiFT. We observe that the model trained on the lowest learning rate of $0.0005$ is the most robust compared to other learning rates we sampled. Comparatively, it attains the second best clean accuracy overall. In~\autoref{fig:ablation-lr-clal}, we see that the degree of double descent for learning rate $0.005$ is severe compared to other values followed by $0.003$. A learning rate of $0.0001$ seems to be too low for the model to converge. In~\autoref{fig:ablation-lr-vraval}, learning rates $0.005$ and $0.003$ lead to the least robust models. 

\begin{figure}[htb]
    \centering
    \begin{subfigure}[b]{0.45\textwidth}
        \centering
        \includegraphics[width=\textwidth]{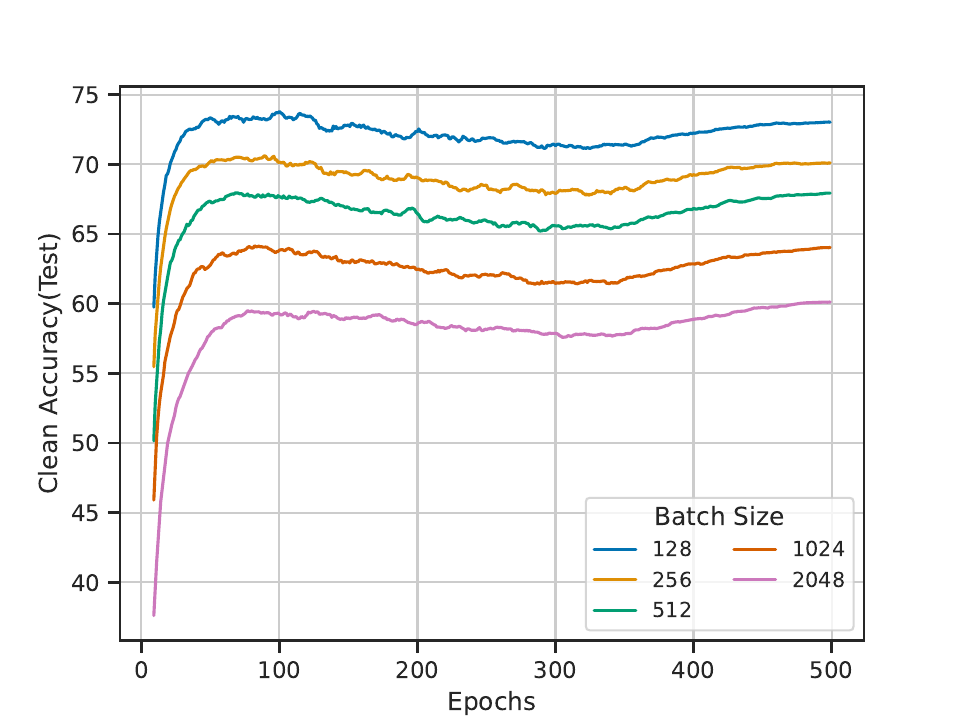}
        \caption{Clean accuracy}
        \label{fig:ablation-bs-clval}
    \end{subfigure}
    \hfill
    \begin{subfigure}[b]{0.45\textwidth}
        \centering
        \includegraphics[width=\textwidth]{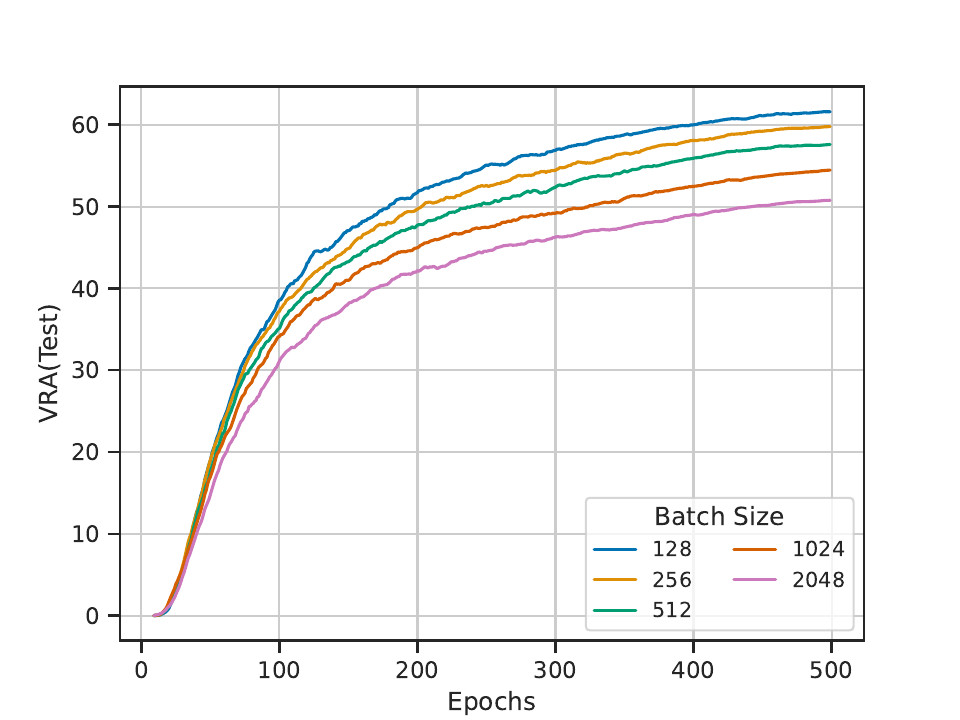}
        \caption{Verified Accuracy}
        \label{fig:ablation-bs-vraval}
    \end{subfigure}

    \caption{Effect of batch size on accuracy.}
    \label{fig:ablation-bs}
\end{figure}

\paragraph{Effect of Batch Size}
We test batch size for the following values: [128, 256, 512, 1024, 2048].~\autoref{fig:ablation-bs} shows the the effect of batch size on clean and verified accuracy. 
Between~\autoref{fig:ablation-bs-clval} and~\autoref{fig:ablation-bs-vraval}, the impact of using a different batch size is more visible compared to the learning rate and dropout ablations. 
This shows that batch size is the more influential in affecting model performance in an adversarial setting. 
We observe that a batch size of $128$ performs the best over the clean and verified accuracy metrics, while $2048$ is the worst performing one. 
This could be tied to the reason that a higher batch size has more hard samples which make it difficult for the the model to generalize. 
A smaller batch size distributes these samples over more number of batches and the model finds it easier to classify them every iteration, hence improving model robustness.

\end{document}

%% file: math_commands.tex
\usepackage{amsmath,amsfonts,bm}

\def\eqref#1{equation~\ref{#1}}

\def\1{\bm{1}}

\DeclareMathAlphabet{\mathsfit}{\encodingdefault}{\sfdefault}{m}{sl}
\SetMathAlphabet{\mathsfit}{bold}{\encodingdefault}{\sfdefault}{bx}{n}

